\documentclass{resart}

\usepackage{multirow}
\usepackage{array}
\usepackage[font=small,labelfont=bf,labelsep=period]{caption}
\usepackage{flushend}
\usepackage{microtype}

\usetimesfont
\atColsBreak{\vskip\baselineskip}

\newenvironment{algorithm}{\vspace{5 pt} \setlength{\baselineskip}{1.1\baselineskip}} {\vspace{5 pt} \par}
\newcommand{\step}[2]{#1. \hspace*{2 pt} \hspace*{#2 em}}

\title{Rapid Feature Extraction for Optical Character Recognition}

\author{M.~Zahid~Hossain\thanks{Department of Computer Science and Electrical Engineering, North South University, Bangladesh. \url{mzhossain@gmx.com}}
\and
M.~Ashraful~Amin\thanks{School of Engineering and Computer Science, Independent University Bangladesh, Bangladesh. \url{aminmdashraful@ieee.org}}
\and
Hong~Yan\thanks{Department of Electronic Engineering, City University of Hong Kong, Hong Kong, China. \url{h.yan@cityu.edu.hk}}
}

\subjects{Pattern Recognition}

\begin{document}
\maketitle

\begin{abstract}
Feature extraction is one of the fundamental problems of character recognition. The performance of character recognition system is depends on proper feature extraction and correct classifier selection. In this article, a rapid feature extraction method is proposed and named as Celled Projection (CP) that compute the projection of each section formed through partitioning an image. The recognition performance of the proposed method is compared with other widely used feature extraction methods that are intensively studied for many different scripts in literature. The experiments have been conducted using Bangla handwritten numerals along with three different well known classifiers which demonstrate comparable results including $94.12\%$ recognition accuracy using celled projection.

\keywords{Character recognition, Feature extraction, Celled projection, Neural network}
\end{abstract}

\section{Introduction}

During the past half century, significant research efforts have been devoted to character recognition to translate human readable characters into machine-readable codes. For Bangla language, it is one of the active research areas waiting for accurate recognition solutions and the accuracy of the recognition solutions is predominantly depends on proper features extraction methods~\cite{TJT96}. There exist many feature extraction methods which have their own advantages or disadvantages over other methods. There are several important criteria of feature extraction methods required to be considered for higher recognition rate. Firstly, an effective feature need to be invariant with respect to character shape variation caused by various writing styles of different individuals and maximize the separability of different character classes. It also needs to represents the raw image data of character through a reduced set of information which are most relevant for classification (i.e., used to distinguish the character classes) to increase the efficiency of classification process. Ease of implementation and fast extraction from raw data are also considered essential for commercial real time applications. Finally, additional preprocessing steps such as noise filtering, binarization, smoothing, thinning reduce the practical efficiency of features. 

Features can be classified into two major categories, statistical and structural features~\cite{MFZ05}. In statistical approach a character image is represented using a set of $n$ features which can be considered as a point in $n$-dimensional feature space. The main goal of feature selection is to construct linear or non-linear decision boundaries in feature space that correctly separate the character images of different classes. Usually statistical approach is used to reduce the dimension of feature set for easy extraction and fast computation where reconstruction of exact original image is not essential. These features are invariant to character deformation and writing style to some extent. Some of the commonly used statistical features for character recognition are projection histograms~\cite{G56}, crossings~\cite{BSG89}, zoning~\cite{B92} and moments~\cite{BJM07} etc.

On the other hand, the structural features such as convex or concave strokes, end points, branches, junctions, connectivity and holes describe the geometrical and topological properties of character. From hierarchical perspective a character is composed of simpler components called primitives~\cite{MFZ05}. In case of structural pattern classification, a character is considered as a combination of primitives and the topological relationship among them. The stroke primitives such as lines and curves construct the structure of a character and generally extracted from skeleton that formed the basic character shape. Usually extraction of structural primitives required various computationally expensive preprocessing including binarization and skeletonization which may cause shape distortions and structural information loss, and as a result character recognition also requires a multilevel complex approximation matching model. However, structural features are more robust against different writing styles and distortions.

\section{Feature Extraction}
\label{sec:feature-extraction}

This section described some of the effective and well studied statistical feature extraction methods from literature for making comparison with the proposed feature.

\subsection{Crossings}

Crossing is one of the popular statistical features for recognizing handwritten character~\cite{BSG89}. It is defined as number of transition from background to foreground or foreground to background along a straight line though out the image. In other word it counts the number of stroke on a line from one side to another side thought the image. In this experiments crossing is computed for every column and row to construct the feature vector of the image. Unlike other features this feature is not influenced by the width of strokes and can be computed without skeletonizing the image.

\subsection{Fourier Transforms}

The Fourier transformation is used in many different ways in character recognition process~\cite{G72}. Transformation of character images the Fourier domain provides valuable information about character structure. The Fourier domain low frequency components denote basic shape and high frequency components denote finer details. For handwritten character recognition process basic shape structure are essential than finer details because finer details highly influence by the noise and writing style. We construct the feature vector for training and classification using $64$ lowest frequency components (to reduce the dimension of feature vector) discarding high frequency components in spectrum. It is observed that the differences of feature vectors among character classes are sometimes small because changes in time domain do not always produce distinguishable changes on the Fourier domain. Thus some classifiers unable to provide higher recognition accuracy.

\subsection{Moments}

Moment invariants are extensively studied as a feature extraction method for image processing and pattern recognition fields~\cite{H62, L92}. There exist different invariants of moments for efficient and effective extraction of features from images of different domains~\cite{BJM07}. Two dimensional moments of order $(p + q)$ of a gray level or binary image can be defined as 
\begin{equation*}
m_{pq} = \sum_{x}\sum_{y}x^{p}y^{q} f(x, y)
\end{equation*}
where $p,q = 0, 1, 2, \ldots, \infty$ and the function $f(x, y)$ provides pixel value of $x$th column and $y$th row of the image. The sums are taken over all the pixels of the image. The central moments with translation invariance of order $(p + q)$ can be written as
\begin{equation*}
\mu_{pq} = \sum_{x}\sum_{y}(x - \bar{x})^{p}(y - \bar{y})^{q} f(x, y)
\end{equation*}
where $\bar{x} = m_{10} / m_{00}$ and $\bar{y} = m_{01} / m_{00}$. The translation invariant central moments place the origin at the center of gravity of the image. In our case, scale invariant moments are not essential because we used normalized images for all our experiments. We construct a feature vector with fifteen translation invariant central moments i.e. $\mu_{00}$, $\mu_{10}$, $\mu_{01}$, $\mu_{11}$, $\mu_{20}$, $\mu_{02}$, $\mu_{22}$, $\mu_{30}$, $\mu_{03}$, $\mu_{21}$, $\mu_{12}$, $\mu_{31}$, $\mu_{13}$, $\mu_{40}$, $\mu_{04}$. We use up to fourth order central moments which is essential for our study because it is observed that higher order moments are sensitive to noise and variation of writing style. Hu~\cite{H62} introduced rotation invariant moments. We also studied the seven Hu moments for our experiments but the recognition rate is poor in compare to other features.

\subsection{Projection Histograms}

Glauberman~\cite{G56} used projection histograms in a hardware based OCR system in 1956. According to this feature, image is scan along a line from one side to another side and number of foreground pixel on the line is counted. Thus it is also known as histogram projection count and can be represented as $H_{i} = \sum_{j} f(i, j)$ for horizontal projection where $f(i, j)$ is the pixel value of $i$th row and $j$th column of the image. Here the background pixel is considered as $0$ and foreground pixel is considered as $1$. Similarly, vertical projection histogram can be calculated. This feature is widely used in several preprocessing steps of document image segmentation where it is used for segmenting text lines, words and characters~\cite{CP98}. In the experiments, we calculate both horizontal and vertical projection histograms and combine them into a feature vector for training and testing. This measurement is not image size invariant but all the character data used for the recognition process have same size. The feature does not consider stroke width variation in handwritten characters.

\subsection{Zoning}

The commercial OCR system named Calera is developed based on zonal feature extraction method which is reported in Bokser~\cite{B92}. According to his study contour extraction and thinning are not reliable for self-touching characters. To extract this feature an image is divided into some non-overlapping or overlapping zones (Cao~\cite{CAS94} studied the overlapping zones viewed as a fuzzy borders around the zones for character image). Then the number of foreground pixel is counted and the density is computed for each zone. Sometimes zoning is considered with other features (e.g., contour direction) but in this text we limit the use of the word zoning only for pixel density feature because it is fast and simple enough to compare with other features used here. Zoning is relatively scaling and slant invariant. The feature vector of the experiments is designed to contain the densities of $4 \times 4 = 16$ zones for each image. We also studied pixel densities of $3 \times 3 = 9$ zones for $15 \times 15$ image size but the recognition rate is lower than that of $16$ zones.

\subsection{Celled Projections}

\begin{figure}[tbh]
\centering
\includegraphics[scale=0.50]{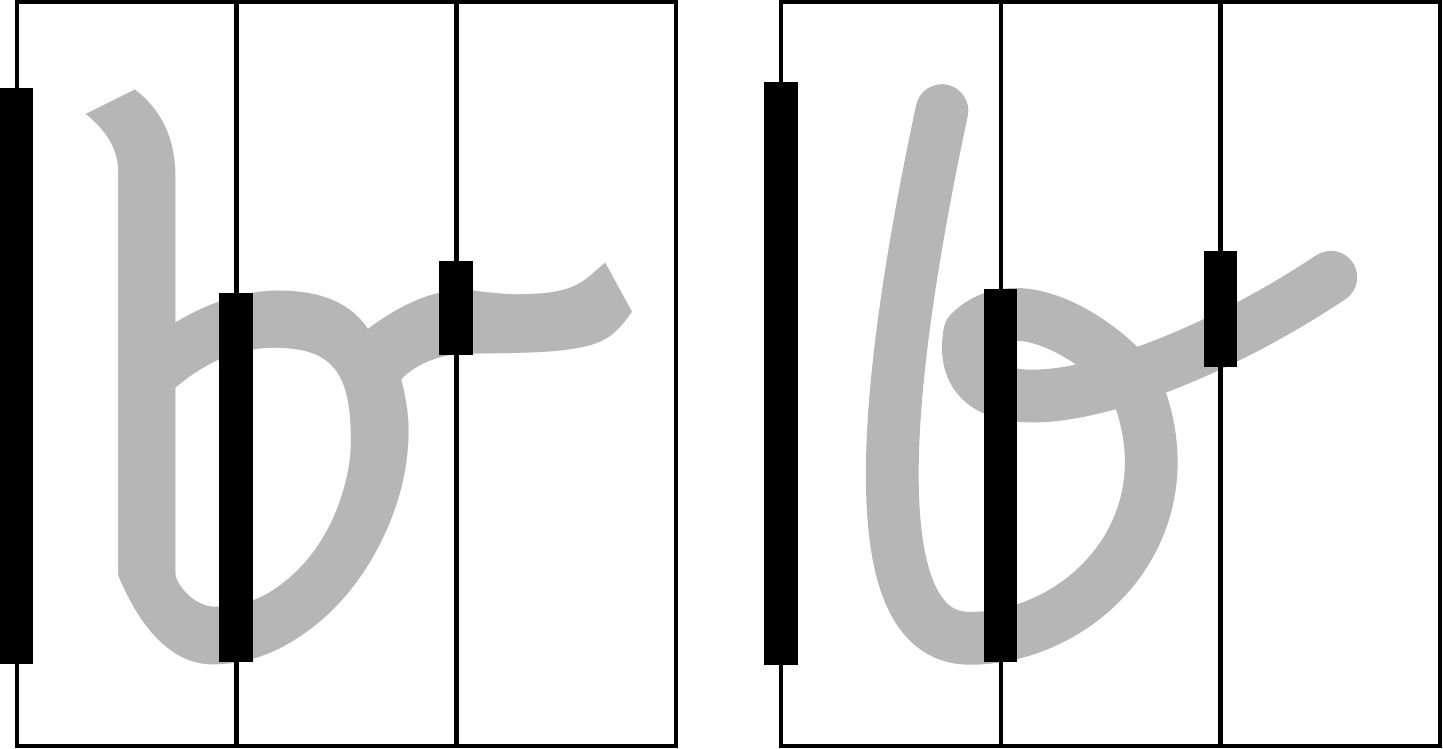}\\
\caption{An example of the celled projection. The geometric shapes on the figure represent Bangla numeral eight in standard form (on left) and handwritten distorted form (on right). It is noticeable that even with those distortions the celled projection of both character are quite similar.}
\label{fig:celled-projection}
\end{figure}

In our proposed feature extraction method of horizontal projections, a character image is partitioned into $k$ regions as shown in Figure~\ref{fig:celled-projection} and then the projection is taken for each region. For horizontal celled projection the feature vector of $r$th cell (or region) of an $m \times n$ image can be written as $P_{r} = \langle p_{1}, p_{2}, \ldots, p_{m} \rangle$ where $p_{i}$ can be formalized as $p_{i} = \bigvee_{j = 1}^{n / k} f(i, \frac{n(r - 1)}{k} + j)$ and $f(x, y)$ is the value of the pixel in $x$th row and $y$th column. Here the background pixel is considered as $0$ and foreground pixel is considered as $1$. The feature vector of the complete image is $V = P_{1} \cup P_{2} \cup \cdots \cup P_{k}$. Using a similar technique vertical and diagonal (or from any other angle) celled projection can be formulated.

\vspace*{5pt}

Although in the algorithm we consider that the input image is a binarized image, it is possible to extract the proposed feature directly from gray scale image using a threshold which separates foreground pixels from background pixels. The arithmetic division operations of the algorithm can be replaced by rearranging the steps with an additional inner loop. The \emph{size} function with an image parameter returns the number of rows and columns of the image. In the algorithm, the \emph{allocate} function reserves a vector in memory of a dimension provided as parameter.

\begin{figure}[tbh]
\begin{algorithm}
$\textsc{Horizontal-Celled-Prjection}(G, k)$\\
\step{01}{0.0} $(m, n) \leftarrow size(G)$\\
\step{02}{0.0} $V \leftarrow allocate(mk)$\\
\step{03}{0.0} $q \leftarrow n / k$\\
\step{04}{0.0} \textbf{for} $i \leftarrow 1$ \textbf{to} $m$\\
\step{05}{1.5} \textbf{for} $j \leftarrow 1$ \textbf{to} $n$\\
\step{06}{3.0} \textbf{if} $G_{i, j} = 1$ \textbf{then}\\
\step{07}{4.5} $V_{i + m \lfloor (j - 1) / q \rfloor} \leftarrow 1$\\
\step{08}{4.5} $j \leftarrow q \lceil j / q \rceil$\\
\step{09}{0.0} \textbf{return} $V$
\end{algorithm}
\caption{The algorithm to compute the horizontal celled projection of the proposed feature. The output feature vector $V$ is the celled projection of input image $G$ divided into $k$ sections.}
\label{algo:celled-projection}
\end{figure}

To calculate the vertical celled projection we need to modify few steps of the above algorithm or transpose the input image. In compare to other feature extraction method this method required a small number of logical and arithmetic operations and only need to consider all the pixels of image in worst case. Each feature in $p_{i}$ required only one bit to store and thus a large number of features can be packed into a single machine word which is significantly reduce the storage requirement of a feature vector. Classification procedure can be also accelerated using proper techniques such as measuring hamming distance between machine words instead of measuring Euclidean distance between bits in character recognition process. The ease of implementation is clear from the algorithm which makes the proposed feature extraction method an attractive solution for hardware implementation. We construct the feature vectors using both horizontal and vertical celled projection of four and eight cells. The distortion for writing style has limited effect on this feature extraction technique.

\section{Classification}

We evaluate the performance of different feature extraction methods using three classifiers, $k$-nearest neighbour rule (KNN), probabilistic neural network (PNN) and feed forward back propagation neural network (FBPN).

\subsection{$k$-Nearest Neighbour}

The $k$-nearest neighbour (KNN) is one of the well known classification techniques. Given an unlabelled test pattern $x$ and a set of $n$ labelled pattern $\{x_{1}, x_{2}, \ldots, x_{n}\}$ form the training set. The task of the classifier is to predict the class label of test pattern $x$ from $P$ predefined classes. The KNN classifier finds the $k$ closest neighbours of $x$ and determines the class label of $x$ using majority voting. Usually KNN classifier applies Euclidean distance as the distance metric. Although KNN is one of the simplest and easy to implement classifier, it can provide competitive result even compare to the sophisticated multilevel training based classifier and it is quite clear from our experiments. The performance of KNN classifier depends on the proper choice of $k$ and the distance metric used to measure the neighbours distances. In our experiments, we use Euclidean distance metric.

\subsection{Probabilistic Neural Network}

Probabilistic neural network (PNN) is widely used as solution of pattern classification problem following an approach developed in statistic called Bayesian classification theory. PNN is a special form of radial basis function network used for classification. It uses a supervised learning model to learn from a training set which is a collection of instances or examples. Each instance has an input vector and an output class. The PNN architecture used in these experiments consists of two layers: radial basis layer, competitive layer. It is part of Matlab neural network toolbox~\cite{NI00} function collection. To prepare a PNN classifier for pattern classification, some training is required for the estimation of probability density function associated with classes. Training process is faster for PNN than other neural network model such as backpropagation and it is also guaranteed to converge to an optimal direction as the size of the representative training set increases.

\newcolumntype{M}[1]{>{\centering}m{#1}}
\newcommand{\tn}{\tabularnewline}
\setlength{\extrarowheight}{1.5pt}

\begin{table*}[tbh]
\centering

\caption{Bangla handwritten numerals recognition results using different feature extraction methods and classifiers. The parameters denote additional configurations about features which are broadly described in Section~\ref{sec:feature-extraction}.}
\label{tab:experiments}

\resizebox{\textwidth}{!}{

\begin{tabular}{|M{2.1cm}|M{2cm}|M{0.95cm}|M{0.95cm}|M{0.95cm}
|M{0.95cm}|M{0.95cm}|M{0.95cm}|M{1cm}|M{1cm}|M{1cm}|}
\hline
\multirow{3}{2.1cm}[-10pt]{\centering Feature} & \multirow{3}{2cm}[-10pt]{\centering Parameter} 
& \multicolumn{3}{m{3.8cm}}{\centering $k$-Nearest Neighbour Classifier} 
& \multicolumn{3}{|m{3.8cm}}{\centering Probabilistic Neural Network} 
& \multicolumn{3}{|m{3.8cm}|}{\centering Feed Forward Back Propagation Neural Network} \tn
\cline{3-11}
&
& \multicolumn{3}{m{3.7cm}}{\centering $k$-Neighbours } 
& \multicolumn{3}{|m{3.7cm}}{\centering Spread Factor} 
& \multicolumn{3}{|m{3.7cm}|}{\centering Hidden Layer Neuron} \tn
\cline{3-11}
&
& 3 & 5 & 7 & 0 - 1 & 1 - 2 & 2 - \footnotesize{$\infty$} & 21-30 & 31-40 & 41-50 \tn
\hline
\multirow{3}{2.1cm}[-12pt]{\centering Celled Projections}
& 4 Horizontal \rule[-7pt]{0pt}{24pt}
&  92.17  &  92.17  &  91.97  &  92.60  &  91.50  &  83.93  &  87.97  &  89.40  &  89.37  \tn
\cline{2-11}
& 8 Horizontal \rule[-7pt]{0pt}{24pt}
&  92.43  &  92.30  &  92.17  &  92.30  &  92.27  &  87.00  &  85.67  &  87.37  &  87.30  \tn
\cline{2-11}
& 4 Horizontal \& 4 Vertical
&  94.10  &  93.93  &  93.73  &  93.87  &  94.12  &  89.63  &  91.20  &  92.03  &  91.93  \tn
\hline
Crossings
& Horizontal \& Vertical
&  85.80  &  85.97  &  86.33  &  86.40  &  84.70  &  76.83  &  84.80  &  85.07  &  85.87  \tn
\hline
Fourier Transforms
& 64 Low Frequency
&  71.80  &  72.87  &  73.30  &  47.33  &  71.13  &  73.23  &  66.73  &  67.00  &  67.67  \tn
\hline
Moments
& 15 Central Moments
&  67.60  &  67.37  &  68.07  &  10.00  &  10.00  &  67.57  &  84.73  &  85.23  &  86.70  \tn
\hline
Projection Histograms
& Horizontal \& Vertical
&  82.33  &  82.37  &  82.77  &  82.10  &  82.93  &  83.30  &  81.90  &  82.47  &  82.57  \tn
\hline
Zoning & $4 \times 4$ 
&  90.30  &  90.27  &  90.27  &  89.80  &  77.03  &  76.33  &  86.77  &  87.63  &  87.73  \tn
\hline
\end{tabular}

} %resizebox

\end{table*}

According to the architecture of PNN used in these experiments, if an input is presented, the first layer computes distances from the input vector to the training input vectors and produces a vector whose elements indicate how close the input is to a training input. The second layer sums these contributions for each class of inputs to produce as its net output a vector of probabilities. Finally, a compete transfer function on the output of the second layer picks the maximum of these probabilities, and produces a $1$ for that class and a $0$ for the other classes. The performance of the PNN depends upon the spread factor. The classifier will act as a nearest neighbour classifier if spread factor is near zero. As spread factor becomes larger the designed network will take into account several nearby design vectors. Some disadvantages of PNN including non-generalized model, large memory requirement and slow classification phase promote other neural network architectures in application fields.

\subsection{Backpropagation}

Different artificial neural networks such as feed forward back propagation neural network (FBPN) demonstrated to be useful in practical applications. Neural network develop its information categorization capabilities through learning process from examples known as training. In this training process the network adjust its weights and biases to perform accurate classification. One of the most common learning method used in this training process called back-propagation (BP). When network is presented with a set of training data the BP algorithm compute the difference between the actual output and desired output and feeding back the error exist in the output and correct the weights and biases that are responsible for the error. In our experiments, we consider a simple multilayer feed forward network with a single hidden layer to compare the performance of several feature extraction methods so that their performance are not shadowed by network performance.

\section{Experimental Results}

We have collected $12000$ Bangla numeral samples from $120$ different writers~\cite{BN10}. Each writer were provided with grid sheet and asked to write Bangla numerals from 0 to 9 in appropriate box of the grid for ten times. Writers were suggested to use all their writing style variations to fill the grid sheet. We use a portion of the total dataset for faster training and testing of the described features. The experiments have been conducted on a dataset of $6000$ Bangla numeral samples for training and an independent dataset portion of $3000$ Bangla numeral samples for testing to calculate the recognition performance. All input numeral images are normalized to size $16 \times 16$ after computing their bounding rectangles. For FBPN the numeral samples used for validation are the $20\%$ of the total number of samples used for training to avoid overfitting. We varied the neurons in hidden layer from $21$ to $50$ and divide the total range into three subranges and report the best result for each subrange in the Table~\ref{tab:experiments}. The neuron number in the output layer is always fixed (i.e., 10 neurons). In compare to other features described in this text the training process of celled projection for four cells with FBPN classifier required only half time on average. The subranges for PNN are not equally allocated throughout the range. Since the recognition accuracy decreases with the increment of spread factor over $3.0$ for most of the features (i.e., for them the minimum spread factor chosen for test over $3.0$ provide the best results) but for Fourier transforms and moments provide best recognition accuracy at $9.0$ and $900.0$ respectively. Since there are infinite real values in each subrange, we choose a number of values for test inside each subrange distributed uniformly throughout the subrange. We report the performance of KNN classifier for $k = 3, 5, \:\text{and}\: 7$ in the Table~\ref{tab:experiments} and all the features provide its highest recognition accuracy for these $k$ values. Unlike celled projection the simple classifiers such as KNN and PNN could not provide acceptable recognition rate for moments feature extraction method and it also required a long training time for the complex FBPN classifier to get an acceptable recognition rate. In these experiments, the highest recognition rate achieved for Bangla numerals is $94.12\%$ using celled projection with four horizontal and vertical cells and PNN classifier. It also provide the highest recognition rate $94.10\%$ for the simplest classifier KNN which implies that celled projections do not need additional supports from complex classifiers. Zoning and crossings also provide good recognition accuracy for different classifiers.

\section{Conclusion}

The main purpose of this experiment is to compare the performances of different feature extraction methods including the proposed method in different classifiers. The proposed method achieved $94.12\%$ recognition accuracy with PNN which is the highest recognition accuracy in our experimental arrangements. Each feature described here performs outstanding in some cases and poor in other cases. Thus the aggregate recognition rate of these individual features and classifiers are not excellent but combining different techniques such as different number of celled projection with multiple classifier systems~\cite{BFR10} could provide excellent recognition rate.

\subsection*{Acknowledgement}
This work is partially supported by Independent University, Bangladesh and a grant from City University  of Hong Kong (Project 9610034).

\vskip-\baselineskip
\end{document}